\pdfoutput=1

\documentclass[11pt]{article}

\usepackage[final]{acl_latex}

\usepackage{times}
\usepackage{latexsym}

\usepackage[T1]{fontenc}

\usepackage[utf8]{inputenc}

\usepackage{microtype}

\usepackage{inconsolata}

\usepackage{graphicx}
\usepackage{amsmath}
\usepackage{algorithm}
\usepackage{algpseudocode}
\usepackage{booktabs}
\usepackage{tcolorbox}
\usepackage{float}
\usepackage{amssymb}
\usepackage{colortbl}
\usepackage{multirow} 
\usepackage{footmisc}

\definecolor{ForestGreen}{RGB}{34,139,34}
\definecolor{BrightRed}{RGB}{220, 20, 60}

\title{CER: Confidence Enhanced Reasoning in LLMs}

\author{
    Ali Razghandi\thanks{\ \ Equal contribution.}, {\bf Seyed Mohammad Hadi Hosseini}\footnotemark[1], {\bf Mahdieh Soleymani Baghshah} \\
    Sharif University of Technology \\
    \texttt{\{ali.razghandi, hadi.hosseini17, soleymani\}@sharif.edu}
}

\begin{document}
\maketitle
\begin{abstract}

Ensuring the reliability of Large Language Models (LLMs) in complex reasoning tasks remains a formidable challenge, particularly in scenarios that demand precise mathematical calculations and knowledge-intensive open-domain generation. In this work, we introduce an uncertainty-aware framework designed to enhance the accuracy of LLM responses by systematically incorporating model confidence at critical decision points. 
We propose an approach that encourages multi-step reasoning in LLMs and quantify the confidence of intermediate answers such as numerical results in mathematical reasoning and proper nouns in open-domain generation. Then, the overall confidence of each reasoning chain is evaluated based on confidence of these critical intermediate steps. Finally, we aggregate the answer of generated response paths in a way that reflects the reliability of each generated content (as opposed to self-consistency in which each generated chain contributes equally to majority voting). We conducted extensive experiments in five datasets, three mathematical datasets and two open-domain datasets, using four LLMs. The results consistently validate the effectiveness of our novel confidence-aggregation method, leading to an accuracy improvement of up to 7.4\% and 5.8\% over baseline approaches in math and open-domain generation tasks, respectively.\footnote{Code is publicly available at \url{https://github.com/sharif-ml-lab/CER}.}


\end{abstract}

\section{Introduction} 
Recently, Large Language Models (LLMs) \cite{dubey2024llama, guo2025deepseek, jiang2023mistral, groeneveld-etal-2024-olmo, achiam2023gpt} have garnered significant attention for their strong performance across diverse reasoning tasks, including arithmetic reasoning and open-domain question answering \cite{wei2022chain, marasovic-etal-2022-shot, zelikman2022star, kojima2022large, yang-etal-2024-large-language-models}. Approaches such as self-consistency \cite{wang2022self} and few-shot prompting \cite{brown2020language} have also been introduced to enhance the reasoning process of these models. However, these approaches have notable limitations. For instance, few-shot prompting relies on carefully curated demonstrations to perform well, and poorly chosen ones can have a reverse effect on performance \cite{halawi2023overthinking}. In addition, the self-consistency method faces challenges in scenarios where generated  paths either (1) produce inconsistent answers that do not include the correct solution or (2) predominantly converge on incorrect results \cite{zhang-etal-2023-sac3, wang2024chain}.

Besides that, human intelligence is uniquely characterized by its ability to express and communicate uncertainty, a critical skill for sound decision-making and effective collaboration \cite{cosmides1996humans}. Similarly, in artificial intelligence, accurate uncertainty estimation is essential for risk assessment, error mitigation, and reliable decision-making \cite{10.5555/3045118.3045290, guo2017calibration, tomani2021towards, fadeeva-etal-2024-fact}. 
To improve the reasoning capabilities of LLMs, it is essential to equip them with mechanisms for effectively quantifying and leveraging uncertainty. 

\begin{figure*}[!t]
	\centering	\includegraphics[width=1.15\textwidth]{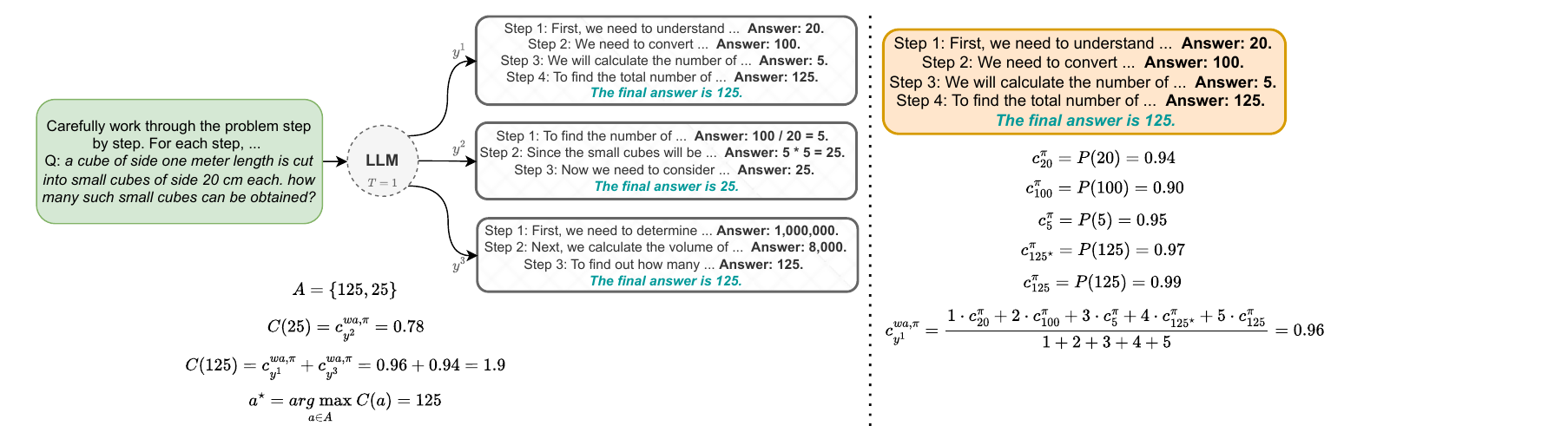}
    
    \caption{\textbf{Illustration of Confidence-Enhanced Reasoning (CER) in LLMs.} On the left, we demonstrate the CER framework. Given an input query, the LLM generates three independent outputs using temperature sampling ($T = 1$). Intermediate answers are bolded, and final answers are highlighted. The confidence of each output is computed, and the most weighted-confident answer—125—is selected. On the right, we illustrate the confidence calculation for the first output. We use multiplication as the step-wise aggregator function (\( f \)) and weighted averaging (\( wa \)) as the path-wise aggregator function (\( g \)). Since the answer 125 appears in both step 4 and the final answer, we mark its first occurrence with * for clarity. The full question and responses from the LLM are provided in Appendix \ref{appendix:F}.}

	\label{fig:main_figure}
\end{figure*}

In this work, we aim to improve reasoning by incorporating uncertainty estimation within a Chain-of-Thought (CoT) process, which consists of a sequence of steps that generate intermediate outputs or answers and ultimately leading to the final answer. At the end of each step, the model is expected to arrive at a certain level of confidence in its output, while some degree of uncertainty is natural throughout a thought due to an incomplete or evolving reasoning step. As a result, we hypothesize that the overall undesired uncertainty of the reasoning chain can be inferred by analyzing the confidence of the tokens that make up the intermediate and final answers. Additionally, these intermediate outputs often exhibit specific characteristics, such as numerical values or proper nouns, that can be readily identified. In fact, we consider these critical tokens in our uncertainty estimation process to enhance the overall accuracy of the reasoning. For mathematical tasks (e.g., GSM8K \cite{cobbe2021gsm8k} ), we prioritize confidence in numerical tokens, while for open-domain generation reasoning (e.g., TriviaQA \cite{joshi-etal-2017-triviaqa}), we focus on the model's confidence in proper nouns (entities, names, locations). 

Based on the above idea, our method comprises three key components: (1) a confidence estimation technique that focuses on evaluating confidence in specific tokens, where a high degree of certainty is crucial, (2) an aggregation strategy for integrating confidence scores across a reasoning chain, and (3) a function that ensembles answers by harnessing the uncertainty within each reasoning chain, resulting in enhanced performance compared to ensemble reasoning methods such as self-consistency.

We evaluated our framework on four LLMs (Llama 3.1, Llama 3.2 \cite{dubey2024llama}, OLMo 2 \cite{groeneveld-etal-2024-olmo}, and Mistral 7B v0.3 \cite{jiang2023mistral}) across five datasets, three mathematical and two open-domain generation benchmarks. Our experiments demonstrate that explicitly incorporating uncertainty in reasoning can enhance accuracy by up to 7.4\% in mathematical tasks and 5.8\% in open-domain question answering. Our contributions are as follows:
\begin{itemize}
    \item 
   By considering the confidence of LLMs in critical points of their responses, we easily compute the uncertainty of an LLM on a generated response that can be useful in aggregating responses generated in multiple chains based on their confidences.

    \item We analyze various functions for each component of our method and identify the best choice to enhance reasoning accuracy.

    \item Empirical validation across various LLMs and benchmarks, showing significant improvements in accuracy without model fine-tuning.
\end{itemize}

\section{Related Work}

\subsection{Reasoning in LLMs}

Recent research has explored various techniques to enhance the reasoning capabilities of LLMs. CoT prompting \cite{brown2020language, kojima2022large} improves multi-step reasoning by generating structured intermediate steps, leading to more transparent and interpretable solutions. Self-consistency \cite{wang2022self} further enhances accuracy by sampling multiple reasoning paths and selecting the most consistent answer. In parallel, question decomposition methods \cite{zhou2022least, dua-etal-2022-successive, khot2022decomposed, ling2023deductive, weng-etal-2023-large} improve coherence by breaking complex queries into simpler sub-questions, though it introduces additional computational overhead. Another promising direction involves search and planning-based methods \cite{wang2023hypothesis, wang2024planning, yao2023tree, besta2024got, xue2025decompose, yang2024buffer}, which systematically explore multiple reasoning trajectories to improve problem-solving. Lastly, integrating external tools—such as web search engines and Python interpreters—extends the model’s capabilities, enabling more precise and efficient task execution across diverse domains \cite{lu2023chameleon, yao2023react, kim2024husky, chen2022program}.
As our approach is grounded in uncertainty estimation, we begin by reviewing existing uncertainty estimation methods, followed by an introduction to uncertainty-aware reasoning techniques, which are the most pertinent to our research.

\subsection{Uncertainty Estimation}
Uncertainty estimation methods can be broadly classified into two categories: black-box \cite{zhang-etal-2023-sac3, xiong2024can, lin2023generating, manakul-etal-2023-selfcheckgpt, chen-mueller-2024-quantifying} and white-box \cite{kuhn2023semantic, duan-etal-2024-shifting, fadeeva-etal-2024-fact, huang2023look} approaches. One approach to uncertainty estimation is training-based confidence estimation \cite{cohen2024don, lin2022teaching, azaria-mitchell-2023-internal}, which improves calibration by incorporating uncertainty estimation directly into the training process. These methods modify the training objective, introduce auxiliary loss functions, or leverage additional supervision to produce more reliable confidence estimates. 
Another approach is verbal-based confidence estimation \cite{tian-etal-2023-just, kadavath2022language}, which prompts the model to explicitly express its confidence through natural language statements. 
Finally, semantic-based uncertainty estimation methods \cite{nikitin2024kernel, kuhn2023semantic, qiu2024semantic, wang2024clue} cluster outputs or reasoning chains that are semantically equivalent, quantifying uncertainty based on the variability of responses within these clusters.

\subsection{Uncertainty-aware reasoning}
An emerging trend leverages uncertainty estimation as a tool to enhance various components of reasoning. One application is in improving few-shot prompting, where uncertainty estimation helps automate the selection of demonstrations \cite{gonen-etal-2023-demystifying, huang2024unlocking, margatina-etal-2023-active}, reducing the need for manually intensive prompt engineering. Another key contribution of uncertainty estimation in reasoning is its role in selecting the most reliable reasoning chain based on confidence \cite{murray-chiang-2018-correcting, kadavath2022language, malinin2020uncertainty}. In such cases, uncertainty acts as a guiding signal, identifying the chain where the model exhibits the highest confidence. Our approach builds on this intuition by enabling a weighted voting mechanism to select the final answer. More importantly, instead of applying our uncertainty estimation function to every token, we focus only on critical tokens, specifically the intermediate answers in a CoT chain.


\section{Confidence Enhanced Reasoning}
Prior research has demonstrated that, analogous to human cognitive processes, enabling LLMs to generate intermediate reasoning steps can substantially enhance their accuracy in complex reasoning tasks. In this work, we aim to extend this approach further by incorporating confidence estimation into the reasoning process. We hypothesize that the final output of each intermediate step—whether a numerical value in mathematical problems or a contextually salient entity in open-domain generative reasoning—serves as a probabilistic signal, providing valuable insight into the model's confidence in that step’s validity. Moreover, these localized confidence scores can be aggregated to estimate the model's overall confidence in the entire reasoning chain. By doing so, we refine the self-consistency voting mechanism: rather than selecting the most frequent answer, we sum the confidence scores of chains arrive at the same conclusion and choose the answer with the highest total confidence.

\subsection{Definitions}
In the following, we present the unified definitions used throughout this paper:

\begin{itemize}
    \item \textbf{Token Probability}: The output probability of token $t$ is derived directly from the model's output logits with a simple softmax function; denoted $p_t$.

    \item \textbf{Word Confidence}: The confidence of a word $w$ generated by the model, calculated using a function $f$ that incorporates all the tokens that make up the word; denoted as 
    \begin{equation}
    c^f_{w}=f(\{p_t|t \in w\}).
    \label{eq:1}
    \end{equation}

    \item \textbf{Path Confidence}: An output sequence generated by the LLM, denoted $y$ and consisting of $n$ steps where $n$ shows the number of the constituent parts of the reasoning paths. Each step is composed of two components: a content and an answer component, denoted as $o$ and $a$, respectively. In our method, the confidence score for each path $y$, obtained by aggregating the confidence values of only the critical points, i.e. the answer components $\{a_j\}_{j=1}^{n}$, on the path through a function $g$ as

    \begin{equation}
    \begin{aligned}
    c^{g, f}_{y} &= g\bigl(c^f_{a_1}, \dots, c^f_{a_n}\bigr).
    \end{aligned}
    \label{eq:3}
    \end{equation}
\end{itemize}

\begin{algorithm}[!ht]
\caption{CER Algorithm}
\label{alg:1}
\begin{algorithmic}[1]
     \Require $x$, $P$, $f$, $g$, $K$, \(T\)
    \Ensure $a^*$

\Statex \hspace*{-\algorithmicindent}\textbf{Description:} Given an input prompt \( x \), the language model \( P \) generates responses. The functions \( f \) and \( g \) represent step-wise and path-wise aggregation, respectively. The temperature parameter is denoted by \(T\), and the ensemble consists of \( K \) generations. The final output, is denoted as \( a^* \). 

  \State $\mathcal{P} \gets \emptyset$
  \State $\{y^i\}_{i=1}^{K} \gets P(y | x, T)$
  \For{$i \gets 1$ to $K$}
    \State $y^i = \left\{ (o^i_{j}, a^i_{j}) \right\}_{j=1}^{n^i}$
    \For{\textbf{each} $a^i_{j}$ in $y^i$}
      \State $c^f_{a^i_{j}} \gets f(a^i_{j})$ \Comment{Eq.~\eqref{eq:1}}
    \EndFor
    \State $c^{g, f}_{y^i} \gets g\bigl(c^f_{a^i_{a_1}}, \dots, c^f_{ a^i_{n^i}}\bigr)$ \Comment{Eq.~~\eqref{eq:3}}
    \State $A^i = a^i_{n^i}$
    \State $\mathcal{P} \gets \mathcal{P} \cup \{(c^{g, f}_{y^i}, A^i)\}$
  \EndFor
  \State $\mathcal{A} \gets \{\,a \mid (c^{g, f}_{y^i}, A^i) \in \mathcal{P}\}$
  \For{\textbf{each} $a \in \mathcal{A}$}
    \State $C(a) \gets \sum_{\substack{i=1}}^{K} c^{g, f}_{y^i}.\mathbb{I}(\{A^i = a\})$
    \Comment{Eq.~\eqref{eq:5}}
  \EndFor
  \State $a^* \gets \arg\max_{a \in \mathcal{A}} C(a)$
  \State \Return $a^*$
\end{algorithmic}
\end{algorithm}

\subsection{Method}
At first, we independently generate $K$ response paths $\{ y^1, y^2, \dots, y^K \}$ from the LLM. Next, we break down each response $y^i$ into $n^i$ constituent steps, extracting the answers at different steps as key elements $\{a^i_j\}_{j=1}^{n^i}$ for constructing our confidence subset.
Specifically, the LLM-produced answer in the final step of the generation process $y^i$, i.e. $a^i_{n^i}$, representing the conclusive answer to the question in this path is denoted as $A^i$.

We can compute the confidence of each answer using the function $f$ as in \eqref{eq:1}.
For instance, if $f$ is a multiplication function and $a^i_j$ consists of $r$ tokens $\{t_1, \dots, t_r\}$, the confidence on this special point can be written as:

\begin{equation}
c^{\prod}_{a^i_j} = \prod_{k=1}^{r} p(t_k).
\label{eq:2}
\end{equation}
One other choice of $f$ is mean entropy which is computed as the average entropy of distributions on all tokens in the word. 
Details about different choices of $f$ and subsequent impact on the results are thoroughly examined in Appendix~\ref{appendix:A} and ~\ref{appendix:D}.

Subsequently,  we aggregated the confidence scores from all steps of a path using the function $g$ as
in \eqref{eq:3}. For the path-wise aggregate function $g$, which aggregates the confidence scores of words, we experimented with several formulations. Our primary aggregation method is: 
\begin{equation}
\begin{aligned}
c^{g,f}_{y^i} = \frac{\sum_{j=1}^{n^i} j \cdot c^{f}_{a^i_j}}{\sum_{j=1}^{n^i} j}.
\end{aligned}
\label{eq:4}
\end{equation}
It assigns higher weights to the steps that are closer to the final answer.
Other aggregation schemes we considered include harmonic mean and different kinds of weighted means which are  introduced and assessed in Appendix~\ref{appendix:B} and Appendix~\ref{appendix:D}.

Once path confidence is determined, we further aggregate the confidence scores of all paths that yield the same $A^i$. The answer with the highest aggregate confidence is then selected.

\begin{equation}
\begin{aligned}
\mathcal{A} &= \{\, a \mid a \in \{A^i\}_{i=1}^K\}\\
C(a) &= \sum_{\substack{i=1}}^{K} c_{y^i
}^{g,f}\times\mathbb{I}(A^i = a)\quad \forall a \in \mathcal{A},\\
a^\ast &= \arg\max_{a \in \mathcal{A}} C(a).
\end{aligned}
\label{eq:5}
\end{equation}

\noindent
where $\mathcal{A}$ is the set of unique final answers among $\{A^i\}_{i=1}^K$. \(C(a)\) is the aggregated confidence score for each unique \(a\). Finally, \(a^\ast\) is the best candidate, chosen by maximizing the confidence score over all \(a \in \mathcal{A}\).

The algorithm~\ref{alg:1} summarizes the complete procedure of our method.

\begin{table*}[ht]
    \centering
    \small
    \renewcommand{\arraystretch}{1.2}
    \setlength{\tabcolsep}{6pt}
    
    \begin{tabular}{lccccccc|c}
        \toprule
        \textbf{Models \& Datasets} & \textbf{Self-Consistency} & \textbf{P(True)} & \textbf{PE} & \textbf{NL} & \textbf{NE} & \textbf{LL} & \textbf{Greedy} & \textbf{CER} \\
        \midrule
        
        \multicolumn{7}{l}{\textbf{LLaMA-3.1-8B}} \\ \midrule
        GSM8K  & 89.6  & 87.6  & 85.2  & 86.2  & 86.2  & 83.8 &  82.8 & \textbf{90.0} \textcolor{ForestGreen}{(+0.4\%)} \\
        
        MATH   & 55.4  & 56.8  & 52.0  & 52.8  & 53.6  & 50.4 & 53.4 & \textbf{58.2} \textcolor{ForestGreen}{(+1.8\%)} \\
        
        MathQA & 63.2  & 65.2  & 64.4  & 65.2  & 61.6 & 65.4 & 60.0 & \textbf{68.2} \textcolor{ForestGreen}{(+2.8\%)} \\

        \textbf{Average} & 69.4  & 69.8  & 67.2 & 68.0  & 67.1  & 66.53 & 65.4 & \textbf{72.1} \textcolor{ForestGreen}{(+2.3\%)} \\
        
        \midrule
        \multicolumn{7}{l}{\textbf{Mistral-7B}} \\ \midrule
        GSM8K  & 62.2  & 46.6  & 55.8  & 59.0  & 60.0 & 55.6 & 44.8 & \textbf{65.2} \textcolor{ForestGreen}{(+3.0\%)} \\
        
        MATH   & 20.4  & 13.6  & 19.0  & 20.2  & 20.0 & 19.6 &  17.0 & \textbf{24.0} \textcolor{ForestGreen}{(+3.6\%)} \\
        
        MathQA & 20.8  & 12.4  & \textbf{22.6}  & 20.0  & 19.4  & \textbf{22.6} & 20.2 & \textbf{22.6} \textcolor{ForestGreen}{(+0.0\%)} \\

        \textbf{Average} & 34.4  & 24.2  & 32.4 & 33.0  & 33.1  & 32.6 & 27.3 & \textbf{37.2} \textcolor{ForestGreen}{(+2.8\%)} \\

        \midrule
        \multicolumn{7}{l}{\textbf{OLMo-2-7B}} \\ \midrule
        GSM8K  & 85.0  & 82.0  & 84.4  & 83.8  & 78.0 & 84.8 & 84.2 & \textbf{88.8} \textcolor{ForestGreen}{(+3.8\%)} \\
        
        MATH   & 42.5 & 40.0  & 41.0  & 40.0  & 39.2 & 42.6 & 37.8 & \textbf{48.0} \textcolor{ForestGreen}{(+5.4\%)} \\
        
        MathQA & 52.0  & 51.8  & 44.8  & 50.0  & 48.8 & 47.4 & 45.2 & \textbf{59.4} \textcolor{ForestGreen}{(+7.4\%)} \\

        \textbf{Average} & 59.8  & 57.9  & 56.7 & 57.9  & 53.3  & 58.2  & 55.73 & \textbf{65.1} \textcolor{ForestGreen}{(+5.3\%)} \\

        \midrule
        \multicolumn{7}{l}{\textbf{LLama-3.3-3B }} \\ \midrule
        GSM8K  & 78.4  & 73.2  & 73.0  & 77.0  & 78.6 & 75.2 & 75.2 & \textbf{82.6} \textcolor{ForestGreen}{(+4\%)} \\
        
        MATH   & 51.2  & 44.2  & 44.0  & 42.6  & 40.0 & 40.2 & 46.4 & \textbf{56.0} \textcolor{ForestGreen}{(+4.8\%)} \\
        
        MathQA & 59.6 & 52.2  & 55.6  & 54.2  & 58.4  & 57.4 & 55.4 & \textbf{62.8} \textcolor{ForestGreen}{(+3.2\%)} \\

        \textbf{Average} & 63.0  & 56.5 & 57.5 & 57.9  & 59.0  & 57.6 & 59.0 & \textbf{67.1} \textcolor{ForestGreen}{(+4.1\%)} \\
        
        \bottomrule
    \end{tabular}
    
    \caption{Accuracy comparison across three mathematical datasets—MATH, MATHQA, and GSM8K—on 500 sampled instances evaluated using various baseline methods and the proposed CER approach. The colored values indicate the improvement or decline compared to the best performance of the baselines for each dataset.}
    \label{tab:mainmath}
\end{table*}

\begin{table*}[ht]
    \centering
    \small
    \renewcommand{\arraystretch}{1.2}
    \setlength{\tabcolsep}{6pt}
    
    \begin{tabular}{lccccccc|c}
        \toprule
        \textbf{Models \& Datasets} & \textbf{Self-Consistency} & \textbf{P(True)} & \textbf{PE} & \textbf{NL} & \textbf{NE} &  \textbf{LL} & \textbf{Greedy} & \textbf{CER} \\
        \midrule
        
        \multicolumn{7}{l}{\textbf{LLaMA-3.1-8B}} \\ \midrule
        
        Trivia QA & 62.2 & 64.8 & 58.0 & 58.0 & 60.2 & 59.4 & 61.8 & \textbf{66.0} \textcolor{ForestGreen}{(+1.2\%)} \\
        
        HotPot QA & 10.2 & \textbf{14.4} & 11.0 & 13.4 & 12.6 & 13.2 & 14.2 & \textbf{14.4} \textcolor{ForestGreen}{(+0.0\%)} \\

        \textbf{Average} & 36.2 & 39.6 & 34.5 & 35.7 & 36.4 & 36.3 & 38.0 & \textbf{40.2} \textcolor{ForestGreen}{(+0.6\%)} \\
        
        \midrule
        \multicolumn{7}{l}{\textbf{Mistral-7B}} \\ \midrule
        Trivia QA & 37.0 & 43.2 & 48.6 & 46.0 & 44.2 & 47.0 &  44.8 & \textbf{54.4} \textcolor{ForestGreen}{(+5.8\%)} \\
        
        HotPot QA & 7.2 & 6.4 & 10.2 & 7.6 & 6.8 & 8.8 & 8.4 & \textbf{10.4} \textcolor{ForestGreen}{(+0.2\%)} \\

        \textbf{Average} & 22.1 & 24.8 & 29.4 & 26.8 & 25.5 & 27.9 & 26.6 & \textbf{32.4} \textcolor{ForestGreen}{(+3.0\%)} \\

        \midrule
        \multicolumn{7}{l}{\textbf{OLMo-2-7B}} \\ \midrule
        Trivia QA & 47.0 & 49.0 & 48.0 & 45.2 & 43 & 46.4 &  48.4 & \textbf{50.8} \textcolor{ForestGreen}{(+1.8\%)} \\
        
        HotPot QA & 8.6 & 8.6 & 8.2 & 8.8 & 7.8 & 8.6 & 8.4 & \textbf{10.6} \textcolor{ForestGreen}{(+1.8\%)} \\

        \textbf{Average} & 27.8 & 28.8 & 28.1 & 27.0 & 25.4 & 27.5 & 28.4 & \textbf{30.7} \textcolor{ForestGreen}{(+1.9\%)} \\

        \midrule
        \multicolumn{7}{l}{\textbf{LLama-3.3-3B}} \\ \midrule
        Trivia QA & 48.8 & 50.8 & 45.0 & 43.4 & 42.4 & 41.4 &  49.4 & \textbf{53.0} \textcolor{ForestGreen}{(+2.2\%)} \\
        
        HotPot QA & 9.0 & 8.4 & 6.4 & 6.8 & 7.8 & 7.4 &  9.0 & \textbf{9.2} \textcolor{ForestGreen}{(+0.2\%)} \\

        \textbf{Average} & 28.9 & 29.6 & 25.7 & 25.1 & 25.1 & 24.4 & 29.0 & \textbf{31.1} \textcolor{ForestGreen}{(+1.5\%)} \\

        \bottomrule
    \end{tabular}
    
    \caption{Accuracy comparison on two open-domain QA datasets—Trivia QA and HotPot QA—using 500 sampled instances. The table presents results across multiple baseline methods alongside the proposed CER method. Colored values represent the performance change compared to the best baseline performance.}
    \label{tab:mainmulti}
\end{table*}

\section{Experiments}
\subsection{Experimental Setup}

In this section, we present the experimental setup used to assess our method and compare it with the other methods. 

\paragraph{Models:} We evaluate our approach on a diverse set of LLMs to capture a wide range of architectures and capabilities. Our primary model is \textbf{Meta Llama 3.1 8B Instruct} \cite{dubey2024llama}, a state-of-the-art open source LLM known for its robust performance. To further support our findings, we also conducted experiments on \textbf{Meta Llama 3.2 3B} \cite{dubey2024llama}, representing a powerful yet compact model. Additional experiments were performed using \textbf{Mistral 7B Instruct} \cite{jiang2023mistral}, a model frequently referenced in recent studies, and \textbf{Olmo 2 7B} \cite{groeneveld-etal-2024-olmo}, which exemplifies the latest mixture of expert architectures.

\paragraph{Datasets and Tasks:} We evaluate our method across two task categories: 1) mathematical reasoning and 2)open-domain question answering. For the mathematical tasks, we utilize the following datasets:
\begin{itemize}
    \item \textbf{GSM8K} \cite{cobbe2021gsm8k}: A widely used benchmark that contains mathematical problems with numerical answers.
    \item \textbf{MATH} \cite{hendrycksmath2021}: A dataset that presents more complex mathematical problems than GSM8K. It consists of two parts: numerical and non-numerical answers. We preprocessed the dataset and filtered out all mathematical questions that yield non-numerical answers.
    \item \textbf{Math QA} \cite{amini-etal-2019-mathqa}: A collection of difficult math problems that do not overlap with the MATH dataset.
\end{itemize}
For open-domain question answering, we utilize the following datasets:
\begin{itemize}
    \item \textbf{TriviaQA} \cite{joshi-etal-2017-triviaqa}: A large-scale dataset containing knowledge-intensive questions sourced from Wikipedia.  
    \item \textbf{HotPotQA} \cite{yang-etal-2018-hotpotqa}: A dataset designed for multi-hop reasoning \cite{yang-etal-2024-large-language-models}, requiring models to synthesize information from multiple documents. We preprocessed the dataset by removing all comparison questions and filtering out open-domain generation questions that do not have a proper noun as their answer.
\end{itemize}
Both of these datasets require comprehensive reasoning and are knowledge-intensive.

\paragraph{Evaluation Metrics:} Given our emphasis on reasoning and verifiable problem solving, we adopt accuracy as the main evaluation metric.

\paragraph{Baselines:} We compare our approach against several baselines that include greedy sampling and self-consistency as baselines and also improved versions of self-consistency by incorporating confidence or uncertainty in their voting phase:
\begin{itemize}
    \item \textbf{Greedy Sampling:} Uses straightforward greedy decoding to generate a single response, serving as a baseline for the model's raw performance.
    \item \textbf{Self-Consistency} \cite{wang2022self}: Aggregates multiple response paths to enhance reasoning accuracy.
    \item \textbf{Token "True" Probability} \cite{kadavath2022language}: Determines the final answer based on the probability assigned to the token ``true''.

    \item \textbf{Log Likelihood (LL)} \cite{murray-chiang-2018-correcting}: Multiply the probabilities of all tokens in a response path.
    
    \item \textbf{Normalized Likelihood (NL)} \cite{murray-chiang-2018-correcting}: A length-normalized variant of log likelihood, computed by dividing the log likelihood by the sequence length.
    
    \item \textbf{Predictive Entropy (PE)} \cite{kadavath2022language}: Computes the mean entropy over all tokens in a response path to assess confidence.
    
    \item \textbf{Normalized Entropy (NE)} \cite{malinin2020uncertainty}: A length-normalized variant of predictive entropy, obtained by dividing the entropy by the sequence length.
    
\end{itemize}

More details on the formulation and the aggregation approach of confidence-based methods are provided in the Appendix~\ref{appendix:G}.

\paragraph{Implementation Details:} 
All methods, except for the simple greedy baseline, utilize temperature sampling with $T=1$ to generate responses. The number of generated paths \(K\) is set to 10, a choice supported by previous research. \cite{zhang-etal-2023-sac3, duan-etal-2024-shifting, qiu2024semantic, fadeeva-etal-2024-fact} We aggregated all response paths based on an exact match of the final answer to the question. Our experiments were conducted on a single A100 80G GPU.
We sample 500 data points from each dataset and evaluate our results on these subsets.
Additional details, including input prompts and sample instances from the datasets, can be found in the Appendix~\ref{appendix:C}.

\subsection{Main Results}
\paragraph{Mathematical Reasoning}
 Table~\ref{tab:mainmath} reports the performance of our models on three mathematical datasets under the CER framework, alongside all baseline methods. Notably, our CER approach consistently surpasses every baseline, with its advantage being particularly marked when applied to smaller, less powerful LLMs. In addition, our method yields more significant relative improvements on more challenging datasets. For instance, Llama 3.1 8B records an average relative gain of 2.3\% across the datasets; Mistral 7B, Olmo 2 7B, and Llama 3.2 3B achieve gains of 0.8\%, 5.3\%, and 4.1\%, respectively. An intriguing observation arises from the results on Llama 3.1 7B—the most potent model in our experiments. Although this model already exhibits strong baseline performance, CER not only boosts its overall results but also delivers particularly significant improvements on the more demanding MATH and Allen AI’s Math QA datasets. By contrast, the performance trend for the Mistral model differs: while it shows consistent improvements across all datasets, the performance gap does not widen as markedly on the more challenging problems. This suggests that while CER can unlock additional reasoning capabilities in models with sufficient capacity, its benefits are limited when the underlying model lacks the capacity to solve the problem entirely.
\paragraph{Knowledge Intensive Reasoning}
 Our CER method outperforms all baselines by a substantial margin for open-domain generation tasks requiring intensive knowledge reasoning. Specifically, it delivers average gains of 0.6\% for Llama 3.1 8B, 3.0\% for Mistral 7B, 1.9\% for Olmo 2 7B, and 1.5\% for Llama 3.1 3B. A notable finding is the relatively poor performance of Llama 3.2 3B compared to the other models. Although Llama 3.2 3B outperforms Mistral 7B on mathematical reasoning tasks by a considerable margin, it falls short on knowledge-intensive tasks. We attribute this discrepancy to the nature of questions in Trivia QA and HotPot QA, which demand that specific knowledge be stored within the model’s parameters. In contrast, mathematical reasoning relies primarily on operational and logical skills. Consequently, even though Llama 3.2 3B is distilled from larger, more capable models, its smaller size means it possesses fewer parameters to encapsulate the extensive knowledge required, leading to its diminished performance on knowledge-intensive tasks.
\paragraph{Results Across Different Models}
 As previously noted, our selection of models aims to demonstrate the performance and versatility of our approach across both smaller models and mixtures of experts—a popular choice in recent research. As illustrated in Tables \ref{tab:mainmath} and \ref{tab:mainmulti}, our framework not only achieves strong results with commonly used models such as Llama 3.1 8B and Mistral 7B, but also shows impressive performance on the compact Llama 3.1 3B and the state-of-the-art open-source MoE model, Olmo 2 7B. In every case, CER outperforms all baseline methods across all datasets.
\subsection{Ablation Studies}
We conducted several ablation studies to further elucidate the contributions of individual components and assess the robustness of our approach.

\begin{figure}
	\centering	\includegraphics[width=0.5\textwidth]{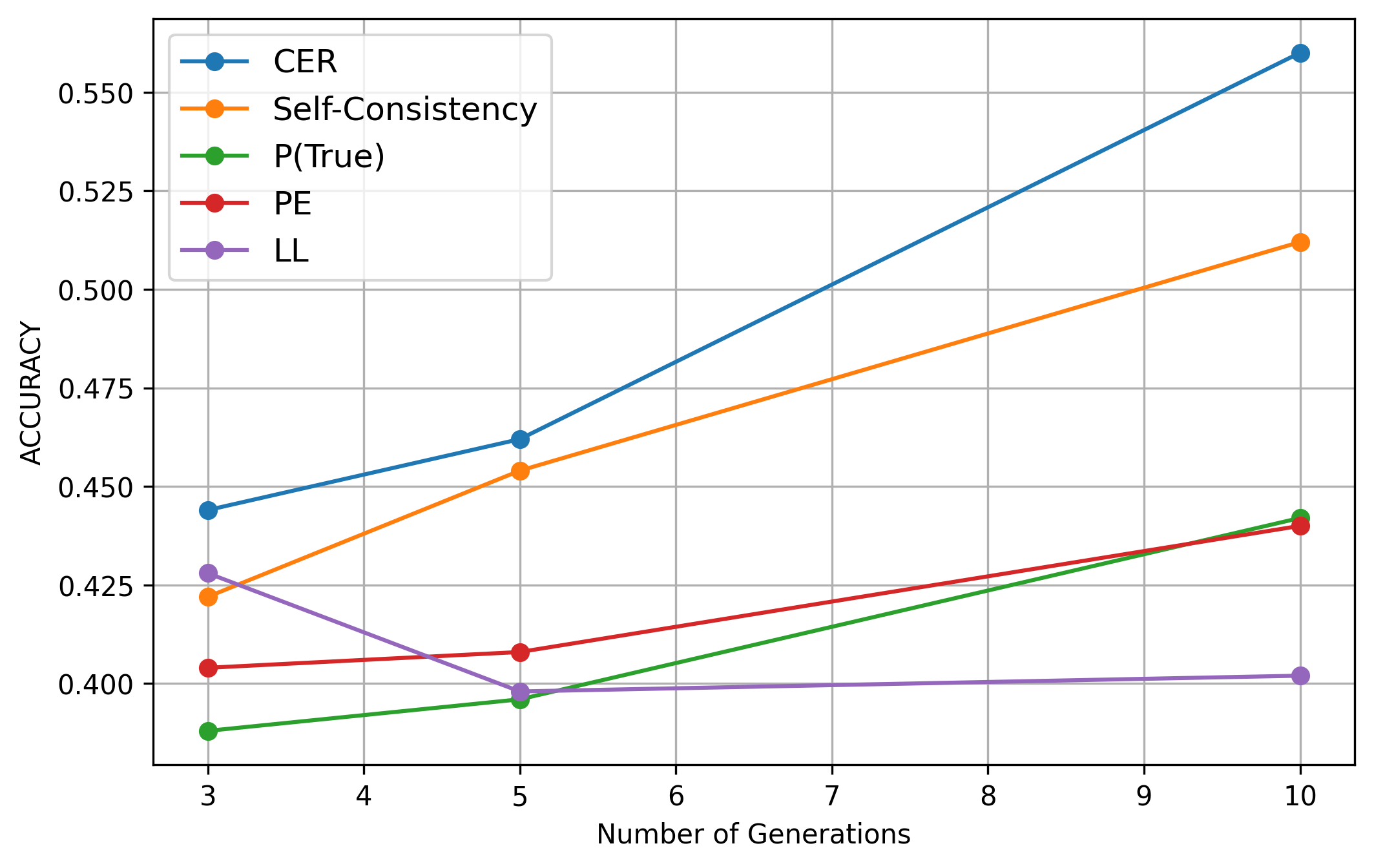}
    
    \caption{Performance comparison of CER and baseline models across different generations $K = \{3, 5, 10\}$ on the LLAMA 3.3-3B model using the MATH dataset.}

	\label{fig:k_generation}
\end{figure}

\begin{figure*}[!t]
	\centering	\includegraphics[width=1\textwidth]{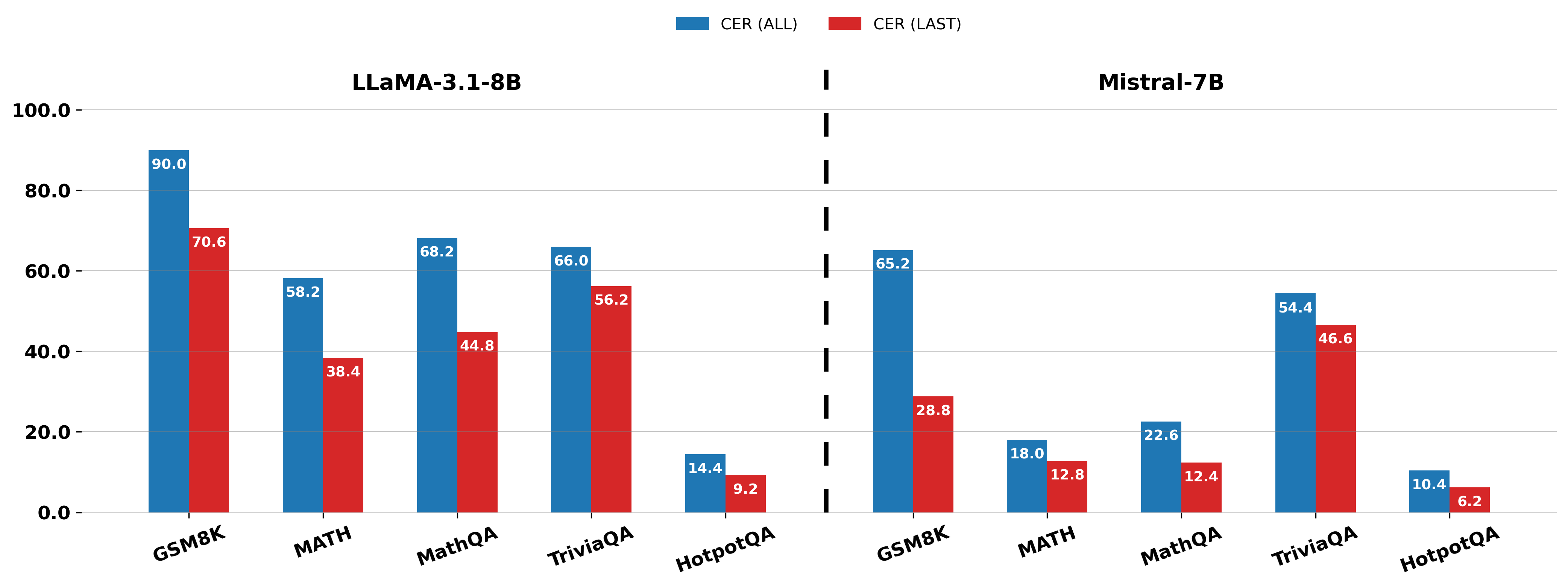}
	\caption{
Ablation study results comparing the performance of the CER method using the last answer confidence (CER-LAST, red) versus the original CER method utilizing all intermediate answers (CER-ALL, blue) across mathematical reasoning datasets (GSM8K, MATH, MathQA) and open-domain question-answering datasets (TriviaQA, HotpotQA). The left side presents results for LLaMA-3.1-8B, while the right side shows results for Mistral-7B. Across all datasets, CER-ALL consistently outperforms CER-LAST, emphasizing the advantage of incorporating intermediate answers for improved accuracy.
	}
	\label{fig:math_multihop_chart}
\end{figure*}

\paragraph{Varying the Number of Paths ($K$)}
 Our first experiment explores the impact of the hyperparameter $K$, which denotes the number of generated paths. As shown in Figure \ref{fig:k_generation}, both CER and all baseline methods benefit from increasing $K$. However, CER consistently outperforms the baselines for every value of $K$.
 
\paragraph{Entropy vs. Probabilities}
 While entropy is commonly used in the literature as a measure of model uncertainty and confidence, we conducted an ablation study comparing the mean entropy over all tokens to the word confidence measure defined in Equation \ref{eq:2}. Appendix~\ref{appendix:D} provides the complete results and the precise formulation of the entropy function, as $f$ is in Appendix~\ref{appendix:A}. 
\paragraph{Different Path-Level Aggregators}
 This study investigated the effect of various path-level aggregator functions, denoted by $g$. Beyond our primary choice of weighted mean aggregation, we experimented with several similar alternatives. The results across these different aggregators were strikingly similar, indicating that the weighted mean is sufficiently effective without requiring further tuning. We also assessed an aggregation function based on the multiplication of word-level confidences along each path, as well as the minimum function—motivated by the adage “a chain is only as strong as its weakest link.” All alternatives yielded comparable results, as detailed in Appendix~\ref{appendix:D}

\paragraph{Last Answer Confidence}
 Finally, we examined the effect of relying solely on the confidence of the last answer to guide the overall reasoning process, thereby excluding intermediate signals. As illustrated in Figure \ref{fig:math_multihop_chart}, this ablation reveals a significant performance gap compared to the original CER method. Although confidence in the final answer is an important indicator, these results confirm that incorporating all intermediate responses leads to superior performance.

\paragraph{Recent Reasoning-based Models}To evaluate appling our method on recent reasoning models, DeepSeek-R1-Distill-Qwen-7B model \cite{guo2025deepseek}, a distilled variant of the state-of-the-art DeepSeek-R1 architecture, is utilized. This model is specifically optimized for mathematical reasoning tasks using the GRPO reinforcement learning method. We compare employing our method against several baseline methods on two benchmarks: GSM8K, which focuses on mathematical reasoning, and TriviaQA, which targets open-domain question answering. The obtained results are shown in Table~\ref{tab:deep_seek}.

 \begin{table*}[ht]
    \centering
    \small
    \renewcommand{\arraystretch}{1.2}
    \setlength{\tabcolsep}{6pt}
    
    \begin{tabular}{lccccccc|c}
        \toprule
        \textbf{Models \& Datasets} & \textbf{Self-Consistency} & \textbf{P(True)} & \textbf{PE} & \textbf{NL} & \textbf{NE} &  \textbf{LL} & \textbf{Greedy} & \textbf{CER} \\
        \midrule
        
        
        GSM8K & 89.8 & 82.6 & 83.8 & 85.2 & 83.8 & 84.0 & 87.2 & \textbf{90.2} \\
        
        Trivia QA & 22.4 & 15.2 & 16.0 & 15.4 & 14.8 & 15.8 & 19.4 & \textbf{24.4} \\
            
        \bottomrule
    \end{tabular}
    \caption{Comparison of CER with baselines on DeepSeek-R1-Distill-Qwen-7B across the GSM8K and TriviaQA datasets.}
    \label{tab:deep_seek}
\end{table*}

\section{Discussion and Conclusion}
In this paper, we introduced a lightweight framework that enhances performance on various reasoning tasks by relying solely on the model's output logits without the need for fine-tuning or task-specific prompts. Our approach bridges the gap between reasoning and uncertainty estimation in LLMs. Through extensive empirical analysis, we observe some key findings: (1) \textbf{Aggregating multiple responses based on the model’s confidence leads to more reliable outcomes.} Unlike self-consistency methods, which generate multiple reasoning paths but assign them equal weight, CER employs a confidence-based approach that assigns weights to each path and aggregates them accordingly. (2) \textbf{Considering intermediate answer tokens to assess confidence is more effective than using all tokens in the chain.} We observe that at the end of each step, the model is expected to arrive at a certain level of confidence in its output, while some degree of uncertainty is natural throughout a thought due to an incomplete or evolving reasoning step. Therefore, a CoT should not be penalized for these natural uncertainties throughout its thoughts. As a result, the overall undesired uncertainty in the reasoning chain is inferred by analyzing the confidence levels of only the critical tokens that constitute the intermediate and final answers, while discarding the confidence of other tokens that may lead to misleading conclusions. 
To achieve this, CER encourages the LLM to express intermediate answers through reasoning and leverages these tokens—like numerical values in mathematical reasoning and proper nouns in open-domain generation. (3) \textbf{Assigning higher weights to reasoning steps closer to the final answer improves performance.} By combining these findings, CER thus enhances the reasoning capabilities of large language models. 

\section {Limitations} 
Our work has several notable limitations. First, the framework has currently been applied only to a limited range of tasks, specifically those that involve mathematical reasoning and knowledge-intensive questions. This restriction arises from the nature of our method, which requires intermediate answers or clear indicators of intermediate results for operation. Additionally, our approach depends on the access to the model output logits. Consequently, our method cannot be employed in scenarios where logits are inaccessible, such as with black-box settings. However, it can be adapted to these settings by modifying the step-wise aggregation, prompting the LLM for intermediate answers, estimating confidence from verbalized confidence at each step, and applying path-wise aggregation to compute overall confidence.
Also, in this study, we focus only on numerical outputs in mathematical reasoning. However, with minor modifications, our approach can also handle non-numerical outputs, such as mathematical proofs. Future research could further extend our framework to mathematical reasoning tasks with non-numerical final answers. 

\section*{Acknowledgments}
We would like to thank everyone who supported us throughout the development of this paper. Special thanks to Dr. Kazemi and Dr. Hassanifard at MCINext for their generous support and for providing the GPU and server resources essential for our research.

\bibliography{acl_latex}

\clearpage

\appendix

\section{Different Choices for step-wise aggregation function ($f$)}
\label{appendix:A}

In this work, we define the stepwise aggregate function (\( f \)) as a function that quantifies the confidence of a word by leveraging the probabilities of its constituent tokens. We consider two common formulations for \( f \):  

\begin{enumerate}  
    \item \textbf{Mean Entropy:} Compute the average entropy of all tokens in a word. This metric represents the confidence of the model when generating the word, where a lower entropy indicates a higher confidence.  
    \item \textbf{Multiplicative Probability:} Determine the overall probability of a word by multiplying the probabilities of its constituent tokens, where a higher value indicates greater confidence.
\end{enumerate}  

\paragraph{Mean Entropy:} 
Let a word \(w\) consist of tokens \{\(t_1, t_2, \dots, t_n\)\} with the corresponding probabilities of the mass functions \(P(T = t_1), P(T = t_2), \dots, P(T = t_n)\) and the corresponding probabilities of the tokens \(p(t_1), p(t_2), \dots, p(t_n)\). We define the mean entropy formulation as follows:

\begin{equation}
\begin{aligned}
f_{\text{entropy}}(w) = -\frac{1}{n} \sum_{i=1}^{n}  H(P(T = t_i))
\end{aligned}
\label{eq:6}
\end{equation}

\begin{equation}
\begin{aligned}
f_{\text{entropy}}(w) = -\frac{1}{n} \sum_{i=1}^{n} \sum_{j} P(t_i = j) \log P(t_i = j)
\end{aligned}
\label{eq:7}
\end{equation}

\paragraph{Multiplicative Probability:} 
Alternatively, the multiplicative probability formulation is given by:

\begin{equation}
\begin{aligned}
f_{\text{mult}}(w) = \prod_{i=1}^{n} p(t_i).
\end{aligned}
\label{eq:8}
\end{equation}

We also performed an ablation study using the mean probability of tokens as an alternative.

\section{Different Choices for path-wise aggregate function ($g$)}
\label{appendix:B}

For the path-wise aggregate function ($g$), which aggregates the confidence scores of words, we experimented with several formulations. Our primary aggregation method is the weighted mean, where \(C_w\) represents the confidence of each word. 
Other aggregation schemes we considered include the following:

For all cases below, let \(\{c_1, \dots, c_n\}\) denote the confidence scores associated with words \(1\) through \(n\).

\begin{itemize}
    \item \textbf{Harmonic Mean:} Aggregates confidences using the harmonic mean. 
    
    \begin{equation}
    \begin{aligned}
    \frac{n}{\frac{1}{c_1} + \dots + \frac{1}{c_n}}
    \end{aligned}
    \label{eq:9}
    \end{equation}
    
    \item \textbf{Weighted Mean:} This approach applies linearly increasing weights to the confidences, based on the intuition that the final steps contribute more to the overall confidence of the path and should therefore receive greater weight.
    \begin{equation}
    \begin{aligned}
     \frac{1 \cdot c_1 + \dots + n \cdot c_n}{1 + \dots + n}
    \end{aligned}
    \label{eq:10}
    \end{equation}

    \item \textbf{Half Split Mean:} A weighted split that assigns half of the total weight to the final answer, with the remaining half distributed uniquely among the other words.
    
    \begin{equation}
    \begin{aligned}
    \frac{1}{2} c_n + \frac{1}{2(n-1)} \sum_{i=1}^{n-1} c_i,\quad n>1.
    \end{aligned}
    \label{eq:11}
    \end{equation}
    
    \item \textbf{Exponential Mean:}  Uses exponents of 2 as the weights to emphasize later steps. 
    
    \begin{equation}
    \begin{aligned}
    \frac{2^0 \cdot c_1 + \dots + 2^{n-1} \cdot c_n}{2^n - 1}
    \end{aligned}
    \label{eq:12}
    \end{equation}

     \item \textbf{Average Log:} Computes the average of the logarithm-transformed confidences. \begin{equation} \begin{aligned} \frac{1}{n} \sum_{i=1}^{n} \log(1+c_i) \end{aligned} \label{eq:13} \end{equation}

    \item \textbf{Minimum:} Uses the minimum confidence among all steps.
    \begin{equation}
    \begin{aligned}
    \min_{i \in \{1,\dots,n\}} c_i
    \end{aligned}
    \label{eq:14}
    \end{equation}

\end{itemize}

Each function represents a distinct hypothesis regarding the relative importance of individual words in the response path. The experimental results comparing these methods are presented in the corresponding section of the paper.

\section{More Implementation Details}
\label{appendix:C}

The prompt is tailored for mathematical reasoning, guiding the LLM through a structured step-by-step process while ensuring it generates an answer at each stage. This is illustrated in Figure \ref{fig:prompt_math_cer}. Similarly, the prompt for open-domain generation is designed to systematically lead the LLM through a logical reasoning process, as shown in Figure \ref{fig:prompt_multihop_cer}.

\begin{figure}[!t]
    \centering
    \begin{tcolorbox}
        \textbf{**Objective**} \\
        Carefully work through the problem step by step. For each step, perform any required reasoning and express the answer at the end of the step. Your response should be in the format Answer: [answer]. After completing the steps, provide the final answer based on the reasoning developed throughout the process. 

        \textbf{**Important Rules**} \\
        1. Perform detailed analyses before concluding the answer. \\
        2. Express intermediate answers explicitly at the end of each step in the format Answer: [answer]. \\
        3. Ensure that your response ends with: \textbf{The final answer is [answer]}, where [answer] is the response to the problem. \\

        Q: \textless question\textgreater
    \end{tcolorbox}
    \caption{Prompt for Math Reasoning}
    \label{fig:prompt_math_cer}
\end{figure}

\begin{figure}[!t]
    \centering
    \begin{tcolorbox}
        \textbf{**Objective**} \\
        Carefully work through the problem step by step, focusing only on the essential steps and limiting your response to five sentences. Your response should end with: The final answer is [answer], where [answer] is the response to the problem.
        \\
        Q: \textless question\textgreater
    \end{tcolorbox}
    \caption{Prompt for Multi-hop Reasoning}
    \label{fig:prompt_multihop_cer}
\end{figure}

\section{More Results}
This section shows the results of the ablation studies for both $f$ and $g$ functions. Table~\ref{tab:f_comparison} shows the results for different choices of $f$, and Table~\ref{tab:g_comparison} shows the results for the $g$ alternatives.

\label{appendix:D}

\begin{table*}[!b]
    \centering
    \begin{tabular}{l c c c c c}
        \toprule
        LLM & \multicolumn{3}{c}{Math datasets} & \multicolumn{2}{c}{Open-domain datasets} \\
        \cmidrule(lr){2-4} \cmidrule(lr){5-6}
        & GSM8K & MATH & MathQA & TriviaQA & HotPotQA \\

        \midrule
        \rowcolor[gray]{0.9}
        \multicolumn{6}{l}{\textbf{Multiplication}} \\
        LLama-3.1-8B & 90.0 & 58.2 & 68.2 & 66.0 & 14.4\\
        Mistral-2-7B & 65.2 & 18.0 & 22.6 & 54.4 & 10.4 \\
        OLMo-2-7B & 88.8 & 48.0 & 59.4 & 50.8 & 10.6 \\
        LLama-3.3-3B & 82.6 & 56.0 & 62.8 & 53.0 & 9.2 \\

        \midrule
        \rowcolor[gray]{0.9} 
        \multicolumn{6}{l}{\textbf{Entropy}} \\
        LLama-3.1-8B & 89.0 & 57.2 & 66.6 & 62.0 & 12.8 \\
        Mistral-2-7B & 65.2 & 21.4 & 22.2 & 52.6 & 9.2 \\
        OLMo-2-7B & 84.0 & 30.0 & 50.0 & 54.0 & 8.0 \\
        LLama-3.3-3B & 85.8 & 50.0 & 60.8 & 50.4 & 8.4 \\
        
        \bottomrule
        
    \end{tabular}
    \caption{Accuracy comparison of different large language models (LLMs) on mathematical reasoning and open-domain question-answering datasets. The models are evaluated on GSM8K, MATH, and MathQA for mathematical reasoning, and TriviaQA and HotPotQA for open-domain tasks. Results are reported for two variations of our step-wise aggregate functions ($f$): Multiplication and Entropy.}
    \label{tab:f_comparison}
\end{table*}

\begin{table*}[!ht]
    \centering
    \begin{tabular}{l c c c c c}
        \toprule
        LLM & \multicolumn{3}{c}{Math datasets} & \multicolumn{2}{c}{Open-domain datasets} \\
        \cmidrule(lr){2-4} \cmidrule(lr){5-6}
        & GSM8K & MATH & MathQA & TriviaQA & HotPotQA \\

        \midrule
        \rowcolor[rgb]{0.8,0.9,1.0}
        \multicolumn{6}{l}{\textbf{Self-Consistency}} \\
        LLama-3.1-8B & 89.6 & 55.4 & 63.2 & 62.2 & 10.2\\
        Mistral-2-7B & 62.2 & 20.4 & 20.8 & 37.0 & 7.2\\
        OLMo-2-7B & 85.0 & 42.5 & 52.0 & 47.0 & 8.6 \\
        LLama-3.3-3B & 78.4 & 51.2 & 59.6 & 48.8 & 9.0 \\
        
        \midrule
        \rowcolor[gray]{0.9} 
        \multicolumn{6}{l}{\textbf{Avg($\log c$)}} \\
        LLama-3.1-8B &  89.6 & 56.8 & 68.6 & 65.8 & 14.6\\
        Mistral-2-7B & 66.8 & 19.2 & 24.8 & 54.2 & 9.8\\
        OLMo-2-7B & 8788.2 & 48.0 & 58.0 & 53.0 & 8.8 \\
        LLama-3.3-3B & 84.0 & 55.6 & 61.6 & 54.8 & 9.6 \\

        \midrule
        \rowcolor[gray]{0.9} 
        \multicolumn{6}{l}{\textbf{min($c$)}} \\
        LLama-3.1-8B & 90.6 & 57.6 & 65.4 & 65.2 & 12.6\\
        Mistral-2-7B & 63.4 & 17.2 & 24.8 & 54.6 & 9.4 \\
        OLMo-2-7B & 89.4 & 46.8 & 58.8 & 49.4 & 8.4 \\
        LLama-3.3-3B & 80.6 & 53.6 & 58.6 & 52.0 & 6.4 \\

        \midrule
        \rowcolor[gray]{0.9} 
        \multicolumn{6}{l}{\textbf{Weighted half}} \\
        LLama-3.1-8B & 89.0 & 59.0 & 68.2 & 67.4 & 13.8 \\
        Mistral-2-7B & 66.8 & 20.0 & 21.6 & 53.2 & 10.0 \\
        OLMo-2-7B & 89.2 & 46.0 & 59.2 & 50.2 & 9.6 \\
        LLama-3.3-3B & 82.2 & 56.2 & 63.2 & 51.0 & 9.6 \\

        \midrule
        \rowcolor[gray]{0.9} 
        \multicolumn{6}{l}{\textbf{Weighted exp-2}} \\
        LLama-3.1-8B & 91.0 & 59.4 & 68.2 & 66.2 & 15.0\\
        Mistral-2-7B &63.4 & 18.2 & 22.6 & 54.0 & 9.2\\
        OLMo-2-7B & 88.8 & 48.2 & 58.4 & 49.6 & 9.0\\
        LLama-3.3-3B  & 82.0 & 55.6 & 62.8 & 55.2 & 8.8 \\

        \midrule
        \rowcolor[gray]{0.9} 
        \multicolumn{6}{l}{\textbf{Harmonic mean}} \\
        LLama-3.1-8B & 90.2 & 56.4 & 65.6 & 66.0 & 12.8 \\
        Mistral-2-7B & 66.4 & 22.4 & 23.2 & 52.0 & 9.4 \\
        OLMo-2-7B & 88.2 & 40.0 & 54.0 & 49.6 & 9.0 \\
        LLama-3.3-3B & 84.2 & 56.0 & 62.6 & 54.0 & 8.8 \\
        \bottomrule
        
    \end{tabular}
    \caption{Accuracy comparison of different variants of our main method $g$ function against the self-consistency baseline on all models, evaluated on mathematical reasoning and open-domain generation datasets.}
    \label{tab:g_comparison}
\end{table*}

\section{Additional Prompt Formats and Evaluation}

We evaluated our method on all models using two additional prompt formats on the GSM8K data set. The second format closely resembles our original prompt but replaces \texttt{Answer:[answer]} with \texttt{Response:[response]} for intermediate steps. The third format offers the most flexibility, allowing the LLM to generate intermediate answers freely at the end of each step without enforcing a specific structure. However, to ensure that the final answer remains easily identifiable and error-free, we still require the model to provide it in a predefined format.

\subsection{Prompt Formats}
The second prompt, which offers greater flexibility than the original, is presented in Figure \ref{fig:second_prompt_cer}, while the third prompt, the most flexible of the three, is shown in Figure \ref{fig:third_prompt_cer}.

\begin{figure}[!t]
    \centering
    \begin{tcolorbox}
            \textbf{**Objective**} \\
            Carefully work through the problem step by step. For each step, perform any required reasoning, and express the response at the end of the step. Your response for each intermediate step should be in the format \texttt{Response: [response]}. After completing the steps, provide the final response based on the reasoning developed throughout the process. \\
                    \textbf{**Important Rules**} \\
            Your response should end with: \textbf{\texttt{The final response is [response]}}, where \texttt{[response]} is the final response to the problem.  \\
            
            Q: \textless question\textgreater
    \end{tcolorbox}
    \caption{Second Prompt (more flexible)}
    \label{fig:second_prompt_cer}
\end{figure}

\begin{figure}[!t]
    \centering
    \begin{tcolorbox}
                \textbf{**Objective**} \\
            Carefully work through the problem step by step. For each step, perform any required reasoning and express the answer at the end of the step. After completing the steps, provide the final answer based on the reasoning developed throughout the process. \\
                                \textbf{**Important Rules**} \\
            Your response should end with \texttt{The final answer is [answer]}, where \texttt{[answer]} is the final response to the problem.  \\
            
            Q: \textless question\textgreater
    \end{tcolorbox}
    \caption{Third Prompt (most flexible)}
    \label{fig:third_prompt_cer}
\end{figure}

\subsection{Prompt Results}
Table \ref{tab:gsm8k_add_prompts} presents the results of different models evaluated on the GSM8K dataset using these prompts, while Table \ref{tab:math_datasets_add_prompts} displays the comparison results of our primary models on the MATH and Allen AI Math datasets on the primary prompt and the most flexible one.

\begin{table*}[!ht]
  \centering
  \begin{tabular}{l c c c}
    \toprule
    \textbf{Model} & \textbf{Original Prompt} & \textbf{Second Prompt} & \textbf{Third Prompt} \\
    \midrule
    Llama 3.1 8B & 90.0 & 88.0 & 90.6 \\
    Mistral 7B   & 65.2 & 62.6 & 67.8 \\
    Llama 3.2 3B & 82.6 & 81.2 & 80.4 \\
    Olmo 2 7B    & 88.8 & 88.4 & 90.2 \\
    \bottomrule
  \end{tabular}
    \caption{Accuracy on GSM8K with different prompt formats}
  \label{tab:gsm8k_add_prompts}
\end{table*}

\begin{table*}[!ht]
  \centering
  \begin{tabular}{l c c | c c}
    \toprule
     & \multicolumn{2}{c}{\textbf{MATH Dataset}} & \multicolumn{2}{c}{\textbf{Allen AI Math Dataset}} \\
    \cmidrule(lr){2-3} \cmidrule(lr){4-5}
    \textbf{Model} & Original Prompt & Third Prompt & Original Prompt & Third Prompt \\
    \midrule
    Llama 3.1 8B & 58.2 & 57.4 & 68.2 & 66.0 \\
    Mistral 7B   & 18.0 & 18.6 & 22.6 & 21.8 \\
    \bottomrule
  \end{tabular}
    \caption{A comparison of accuracy on the MATH and Allen AI Math datasets using the main prompt (Figure~\ref{fig:prompt_math_cer}) and the most flexible prompt (Figure~\ref{fig:third_prompt_cer}).}
      \label{tab:math_datasets_add_prompts}

\end{table*}

\section{Further Exploration of the Dataset}
\label{appendix:E}
Here, we show a sample from each dataset. Table~\ref{tab:dataset_math_cer} shows the mathematical datasets samples and Table~\ref{tab:dataset_multihop_cer} shows the open domain generation QA datasets samples.

\begin{table*}[htb]
    \begin{tabular*}{\textwidth}
    {|p{5.2cm}|p{10cm}|}
        \hline
        \textbf{Question} & \textbf{Sampled Path}\\
        \hline
        \textbf{[GSM8K]} The basketball team went to the steakhouse to eat dinner.  The first player ate a 6-ounce steak.  The second player ate beef tips, containing 8 beef tips, each an ounce in size.  The third player ate a one-pound steak.  And the fourth and fifth players ordered vegetarian meals.  In total, how many ounces of meat were consumed by the team? & 
        \textbf{Step 1}: Calculate the amount of meat eaten by the first player. The first player ate a 6-ounce steak. 0.8 pounds is equal to 6 ounces (using the conversion factor 1 pound is equal to 16 ounces.) \textcolor{blue}{Answer: 6.} \textbf{Step 2}: Calculate the amount of meat eaten by the second player. The second player ate beef tips, containing 8 beef tips, each an ounce in size. Multiply the number of tips (8) by the size of each tip (1 ounce to find the total amount eaten by the second player). 8 * 1 ounce = 8 ounces. \textcolor{blue}{Answer: 8.}        
        \textbf{Step 3}: Calculate the amount of meat eaten by the third player. The third player ate a 1-pound steak. First, we need to convert the 1-pound steak to ounces using the conversion factor 1 pound is equal to 16 ounces. 1 pound multiplied by 16 ounces (pound-ounces) is 16 ounces. \textcolor{blue}{Answer: 16.}
        \textbf{Step 4}: Calculate the total amount of meat eaten by the team by summing the amounts for all the players. Since the fourth and fifth players ate vegetarian meals, they did not eat any meat, so we will only sum the amounts eaten by the first three players. 6 + 8 + 16 = 30 ounces. \textcolor{blue}{Answer: 30.} 
        \textcolor{ForestGreen}{The final answer is 30.} \\
        \hline
        
        \textbf{[MATH]} Find the value of $x$ such that $\sqrt{x - 4} = 4$. & 
        \textbf{Step 1}: To start solving this equation, I will square both sides of the equation to eliminate the square root. By doing this, I will be able to get rid of the radical sign and simplify the equation. \textcolor{blue}{Answer: $(\sqrt{x - 4})^2 = 4^2$} \textbf{Step 2}: The left-hand side of the equation becomes $x - 4$ after squaring the square root because $(\sqrt{y})^2 = y$. The right-hand side of the equation becomes $16$ since $4^2 = 4 \times 4$. \textcolor{blue} {Answer: $x - 4 = 16$.} \textbf{Step 3}: Now, I will isolate $x$ by adding $4$ to both sides of the equation. This gets rid of the $-4$ on the left-hand side of the equation. \textcolor{blue}{Answer: $x - 4 + 4 = 16 + 4$.} \textbf{Step 4}: The left-hand side of the equation simplifies to $x$ since any number minus itself is equal to zero, and adding zero to a number doesn't change its value. The right-hand side of the equation becomes $20$ since $16 + 4 = 20$. \textcolor{blue} {Answer: $x = 20$.} \textcolor{ForestGreen} {The final answer is $20$.} \\
        \hline

        \textbf{[MathQA]} Find a sum for first 8 prime numbers? & 
        \textbf{Step 1}:  To find the sum of the first 8 prime numbers, we need to identify the first 8 prime numbers. \textbf{Answer}: \textcolor{blue} {2, 3, 5, 7, 11, 13, 17, and 19.} \textbf{Step 2}:  Now, let's add these numbers together to get the sum. \textcolor{blue} {Answer: 2 + 3 + 5 + 7 + 11 + 13 + 17 + 19.} \textbf{Step 3}:  Performing the addition: Sum = 2 + 3 = 5, Sum = 5 + 5 = 10, Sum = 10 + 7 = 17, Sum = 17 + 11 = 28  - Sum = 28 + 13 = 41, Sum = 41 + 17 = 58, Sum = 58 + 19 = 77 \textcolor{blue} {Answer: 77} \textcolor{ForestGreen} {The final answer is 77.} \\
        \hline
        
    \end{tabular*}

    \caption{We domenstrate the performance of our method, CER, along with a sampled response path for solving each question in the math dataset. Intermediate answers at each step are highlighted in blue, while the final answer is emphasized in green.}
    \label{tab:dataset_math_cer}
\end{table*}

\begin{table*}[!htb]
    \begin{tabular*}{\textwidth}
    {|p{5.2cm}|p{10cm}|}
        \hline
        \textbf{Question} & \textbf{Sampled Path}\\
        \hline
        \textbf{[TriviaQA]} In which town or city would you find the 'Caledonian Brewery'? & 
        [\textbf{Step 1}] To find the location of the \textcolor{blue}{'Caledonian Brewery'}, let's start by understanding the name itself. [\textbf{Step 2}] 'Caledonian' refers to Caledonia, an ancient Roman name for \textcolor{blue}{Scotland}. \textbf{[Step 3]} The brewery is likely to be in \textcolor{blue}{Scotland} given its name. [\textbf{Step 4}] Caledonian Brewery is specifically located in \textcolor{blue}{Edinburgh, Scotland}. [\textbf{Step 5}] Edinburgh is the capital city of \textcolor{blue}{Scotland}. \textcolor{ForestGreen}{The final answer is Edinburgh.} \\
        \hline
        
        \textbf{[HotPotQA]} The Mutlu Onaral album "Livin\' It" was produced by the bassist for what duo? &  [\textbf{Step 1}] To find the answer, we need to look for information about the Mutlu Onaral album \textcolor{blue} {"Livin\' It"} and find a connection to a duo. [\textbf{Step 2}] After researching "Livin\' It," I found it\'s by Mutlu Onaral, a Turkish-born singer-songwriter known for being a member of \textcolor{blue}{Hall \& Oates}. [\textbf{Step 3}] The American duo Hall \& Oates consists of \textcolor{blue}{Daryl Hall and John Oates}. [\textbf{Step 4}] The bassist for Hall \& Oates, which is none other than bassist and vocalist \textcolor{blue}{Klyde Jones} is the lead vocalist who is also the bassist. \textcolor{ForestGreen} {The final answer is Hall \& Oates}. \\
        \hline
        
    \end{tabular*}

   \caption{We demonstrate the performance of our method, CER, along with a sampled response path for solving each question in the open-domain generation dataset. Each step is clearly separated by a period and labeled as \textbf{[Step i]}, where \textit{i} represents the step number. Intermediate answers, including the last proper noun in each step, are highlighted in blue, while the final answer is emphasized in green.}

    \label{tab:dataset_multihop_cer}
\end{table*}

\section{Full question and responses related to the main figure}
\label{appendix:F}
Table~\ref{tab:answer_samples} shows a question as an input sample to the LLm with our prompt and the corresponding generated paths and their intermediate steps. 

\begin{table*}[!t]
    \centering
    \begin{tabular*}{\textwidth}{|p{5.2cm}|p{10cm}|}
        \hline
        \textbf{Question} & \textbf{Sampled Paths}\\
        \hline
        \multirow{3}{5.2cm}{Carefully work through the problem step by step. For each step, perform any required reasoning, and express the answer at the end of the step, Your response should be in the format Answer: [answer]. After completing the steps, provide the final answer based on the reasoning developed throughout the process. Your response should end with The final answer is [answer], where [answer] is the response to the problem. Q: a cube of side one meter length is cut into small cubes of side 20 cm each. how many such small cubes can be obtained?
        } &  \textbf{Step 1}: First, we need to understand the size of the large cube and the size of the small cubes. The large cube has a side length of 1 meter, and the small cubes have a side length of 20 cm.  \textcolor{blue}{Answer: 20.} \textbf{Step 2}: We need to convert the side length of the large cube from meters to centimeters for a uniform unit of measurement. Since 1 meter = 100 centimeters, the side length of the large cube is 100 cm.  \textcolor{blue}{Answer: 100.} \textbf{Step 3}: We will calculate the number of small cubes that can fit along the side of the large cube. The number of small cubes on one side of the large cube is 100 cm / 20 cm = 5. \textcolor{blue}{Answer: 5.} \textbf{Step 4}: To find the total number of small cubes, we will multiply the number of small cubes along the length, width, and height. So, the total number of small cubes = 5 x 5 x 5 = 125.  \textcolor{blue}{Answer: 125.} \textcolor{ForestGreen}{The final answer is 125.} \\  
        \cline{2-2}
        & \textbf{Step 1}: To find the number of small cubes in one dimension, we will divide the length of the big cube in meters by the side length of the small cube in meters. Since 1 meter = 100 cm, and small cube side = 20 cm,  \textcolor{blue}{Answer: 100 / 20 = 5}.  \textbf{Step 2}: Since the small cubes will be cubical in shape, we need to get the result for all the dimensions (length, breadth, and height) and then multiply them to get the total number of cubes.  We've already found out that 5 cubes can fit along the length and breadth (in step 1). As the cube is cube-shaped, the same number of cubes will fit along the height. \textcolor{blue}{Answer: 5 * 5 = 25.}  \textbf{Step 3}: Now we need to consider all the combinations (length, breadth, and height) to get the final answer. But since we already got the same result for length and breadth, there's no need to multiply again. \textcolor{blue}{Answer: 25.} \textcolor{ForestGreen}{The final answer is 25.} \\  
        \cline{2-2}
        &  \textbf{Step 1}: First, we need to determine the volume of the large cube. Since the side length of the large cube is 1 meter, which is equivalent to 100 centimeters, we calculate its volume.  Volume of the large cube = side\textasciicircum3 = 100\textasciicircum3 = 1,000,000 cubic centimeters. \textcolor{blue}{Answer: 1,000,000.} \textbf{Step 2}: Next, we calculate the volume of a small cube. The side length of a small cube is given as 20 cm, so its volume is calculated by cubing this side length. Volume of a small cube = side\textasciicircum3 = 20\textasciicircum3 = 8,000 cubic centimeters. \textcolor{blue}{Answer: 8,000.} \textbf{Step 3}: To find out how many small cubes can be obtained from the large cube, we divide the volume of the large cube by the volume of one small cube.  Number of small cubes = Volume of the large cube / Volume of a small cube = 1,000,000 / 8,000 = 125. \textcolor{blue}{Answer: 125.}  \textcolor{ForestGreen}{The final answer is 125.} \\  
        \hline
    \end{tabular*}

    \caption{We present the complete prompt and full answers for our main figure. This table includes the full prompt along with three sample paths, where intermediate answers at each step are highlighted in blue, and the final answer is emphasized in green.}
    \label{tab:answer_samples}
\end{table*}

\section{Examination of Confidence Baselines}
\label{appendix:G}

We examine baseline methods that integrate confidence measures, specifically Log-Likelihood (LL), Normalized-Length Likelihood (NL), Predictive Entropy (PE), and Normalized-Length Entropy (NE). Let $ P_{\theta} $ represent the LLM, and denote $ N $ as the number of generated tokens, expressed as $ \{y_1, \dots, y_N\} $.

\begin{itemize}
    \item \textbf{Log-Likelihood (LL):} Computes the likelihood of a response path by multiplying the probabilities of all tokens in the sequence. After evaluating the confidence of each response, the answer with the highest confidence—or equivalently, the one with the lowest negative log-likelihood—is selected. Its corresponding equation is:
    \begin{equation}
        \text{LL} = -\sum_{t=1}^{N} \log P_{\theta}(y_t \mid y_{1:t-1}, x)
        \label{eq:15}
    \end{equation}
    
    \item \textbf{Normalized Likelihood (NL):} Computes a normalized version of the log-likelihood for a response path by multiplying the probabilities of all tokens in the sequence and normalizing the value by the length of the generated response ($N$). The answer with the highest confidence—or equivalently, the one with the lowest negative normalized-length likelihood—is selected. Its corresponding equation is:
    \begin{equation}
        \text{NL} = \frac{-1}{N} \sum_{t=1}^{N} \log P_{\theta}(y_t \mid y_{1:t-1}, x)
        \label{eq:16}
    \end{equation}
    
    \item \textbf{Predictive Entropy (PE):} Computes the mean entropy over all tokens in a response path to assess confidence. The answer with the highest confidence—or equivalently, the one with the lowest predictive entropy—is selected. Its corresponding equation is:

    \begin{multline}
\text{PE} = -\sum_{t=1}^{N} P_{\theta}(y_t \mid y_{1:t-1}, x) \\
\quad \cdot \log P_{\theta}(y_t \mid y_{1:t-1}, x)
\label{eq:17}
\end{multline}

    \item \textbf{Normalized Entropy (NE):} A normalized version of predictive entropy that accounts for sequence length. The answer with the highest confidence—or equivalently, the one with the lowest normalized entropy—is selected. Its corresponding equation is:
    \begin{multline} \text{NE} = \frac{-1}{N} \sum_{t=1}^{N} P_{\theta}(y_t \mid y_{1:t-1}, x) \\ \quad \cdot \log P_{\theta}(y_t \mid y_{1:t-1}, x) \end{multline}
\end{itemize}

\end{document}